\documentclass[journal]{IEEEtran}
\usepackage{subcaption}
\usepackage{mathtools}
\usepackage{amssymb}
\usepackage{booktabs}
\usepackage{comment}

\DeclareMathOperator*{\argmin}{arg\,min}
\ifCLASSINFOpdf
   \usepackage[pdftex]{graphicx}
  \usepackage{xcolor}

   \graphicspath{Fig/}

\else

  \usepackage{here}
  \usepackage[dvips]{graphicx}
  \usepackage[driver]{graphicx}
  \usepackage{epstopdf}
  \AppendGraphicsExtensions{.tif}
  \graphicspath{Fig/}

\fi
\usepackage{graphicx}

\usepackage{tabularx}
\usepackage{amsmath}
\usepackage{amsfonts}
\usepackage{amsthm}
\usepackage{bm}
\usepackage{mathtools}

\usepackage{algorithm,algpseudocode}

\usepackage{enumitem}

\theoremstyle{definition}

\title{Noise Learning Based Denoising Autoencoder}

\usepackage{graphicx}

\author{Woong-Hee Lee,~%~\IEEEmembership{Member,~IEEE,}
        Mustafa~Ozger,~%~\IEEEmembership{Fellow,~OSA,}
        Ursula~Challita,~%~\IEEEmembership{Fellow,~OSA,}
        and~Ki~Won~Sung,~\IEEEmembership{Member,~IEEE}%,~\IEEEmembership{Life~Fellow,~IEEE}% <-this % stops a space
\thanks{This work was supported by a Korea University Grant. This research was supported by the BK21 FOUR(Fostering Outstanding Universities for Research) funded by the Ministry of Education(MOE, Korea) and National Research Foundation of Korea(NRF). This work was partly funded by the European Union Horizon 2020 Research and Innovation Programme under the EU/KR PriMO-5G project with grant agreement No 815191. \textit{(Corresponding author: Ki Won Sung.)}}
\thanks{W.-H. Lee is with the Department of Control and Instrumentation Engineering, Korea University, Republic of Korea (e-mail: woongheelee@korea.ac.kr).}
\thanks{U. Challita is with Ericsson Research, Stockholm, Sweden
(e-mail: ursula.challita@ericsson.com).}
\thanks{M. Ozger and K. W. Sung are with the School of Electrical Engineering and Computer Science, KTH Royal Institute of Technology, Stockholm, Sweden (e-mail:\{ozger and sungkw\}@kth.se).}
}

\begin{document}

%\markboth{Submitted for publication,~October~2019}%

\maketitle

\begin{abstract}
This letter introduces a new denoiser that modifies the structure of denoising autoencoder (DAE), namely noise learning based DAE (nlDAE). The proposed nlDAE learns the noise of the input data. Then, the denoising is performed by subtracting the regenerated noise from the noisy input. Hence, nlDAE is more effective than DAE when the noise is simpler to regenerate than the original data. To validate the performance of nlDAE, we provide three case studies: signal restoration, symbol demodulation, and precise localization. Numerical results suggest that nlDAE requires smaller latent space dimension and smaller training dataset compared to DAE.
\end{abstract}

\IEEEpeerreviewmaketitle

\begin{IEEEkeywords}
machine learning, noise learning based denoising autoencoder, signal restoration, symbol demodulation, precise localization.
\end{IEEEkeywords}

\section{Introduction}

Machine learning (ML) has recently received much attention as a key enabler for future wireless communications \cite{challita2020machine,chen2019artificial,Amin_ML}.
While the major research effort has been put to deep neural networks, there are enormous number of Internet of Things (IoT) devices that are severely constrained on the computational power and memory size. Therefore, the implementation of efficient ML algorithms is an important challenge for IoT devices, as they are energy and memory limited.
Denoising autoencoder (DAE) is a promising technique to improve the performance of IoT applications by denoising the observed data that consists of the original data and the noise \cite{sun2019application}.
DAE is a neural network model for the construction of the learned representations robust to an addition of noise to the input samples {\cite{vincent2008extracting, bengio2013generalized}}.
The representative feature of DAE is that the dimension of the latent space is smaller than the size of the input vector. It means that the neural network model is capable of encoding and decoding through a smaller dimension where the data can be represented.

The main contribution of this letter is to improve the efficiency and performance of DAE with a modification of its structure.
Consider a noisy observation $Y$ which consists of the original data $X$ and the noise $N$, i.e., $Y = X+N$.
From the information theoretical perspective, DAE attempts to minimize the expected reconstruction error by maximizing a lower bound on mutual information $I(X;Y)$.
In other words, $Y$ should capture the information of $X$ as much as possible although $Y$ is a function of the noisy input.
Additionally, from the manifold learning perspective, DAE can be seen as a way to find a manifold where $Y$ represents the data into a low dimensional latent space corresponding to $X$.
However, we often face the problem that the stochastic feature of $X$ to be restored is too complex to regenerate or represent. This is called the curse of dimensionality, i.e., the dimension of latent space for $X$ is still too high in many cases.

What can we do if $N$ is simpler to regenerate than $X$? It will be more effective to learn $N$ and subtract it from $Y$ instead of learning $X$ directly.
In this light, we propose a new denoising framework, named as noise learning based DAE (nlDAE).
The main advantage of nlDAE is that it can maximize the efficiency of the ML approach (e.g., the required dimension of the latent space or size of training dataset) for capability-constrained devices, e.g., IoT, where $N$ is typically easier to regenerate than $X$ owing to their stochastic characteristics.
To verify the advantage of nlDAE over the conventional DAE, we provide three practical applications as case studies: signal restoration, symbol demodulation, and precise localization.

The following notations will be used throughout this letter.
\begin{itemize}
    \item $\text{Ber}, \text{Exp}, \mathcal{U}, \mathcal{N}, \mathcal{CN}$: the Bernoulli, exponential, uniform, normal, and complex normal distributions, respectively.
    \item $\mathbf{x},\mathbf{n},\mathbf{y} \in \mathbb{R}^P$: the realization vectors of random variables $X,N,Y$, respectively, whose dimensions are $P$.
    \item $P'(<P)$: the dimension of the latent space.
    \item $\mathbf{W} \in \mathbb{R}^{P' \times P}, \mathbf{W}' \in \mathbb{R}^{P \times P'}$: the weight matrices for encoding and decoding, respectively.
    \item $\mathbf{b} \in \mathbb{R}^{P'}, \mathbf{b}' \in \mathbb{R}^{P}$: the bias vectors for encoding and decoding, respectively.
    \item {$\mathcal{S}$: the sigmoid function, acting as an activation function for neural networks, i.e., $\mathcal{S}(a) = \frac{1}{1+e^{-a}}$, and $\mathcal{S}(\mathbf{a})=(\mathcal{S}(\mathbf{a}[1]),\cdots, \mathcal{S}(\mathbf{a}[P]))^T$ where $\mathbf{a}\in \mathbb{R}^P$ is an arbitrary input vector.}
    \item $f_{\theta}$: the encoding function where the parameter $\theta$ is $\{ \mathbf{W}, \mathbf{b} \}$, i.e., $f_{\theta}(\mathbf{y}) = \mathcal{S}(\mathbf{W}\mathbf{y}+\mathbf{b})$.
    \item $g_{\theta^{'}}$: the decoding function where the parameter $\theta^{'}$ is $\{ \mathbf{W}', \mathbf{b}' \}$, i.e., $g_{\theta^{'}}(f_{\theta}(\mathbf{y})) = \mathcal{S}(\mathbf{W}'f_{\theta}(\mathbf{y})+\mathbf{b}')$.
    \item $M$: the size of training dataset.
    \item $L$: the size of test dataset.
\end{itemize}

\section{Method of nlDAE}

\begin{figure*}
     \centering
     \begin{subfigure}[b]{0.27\textwidth} %41
         \centering
         \includegraphics[width=\textwidth]{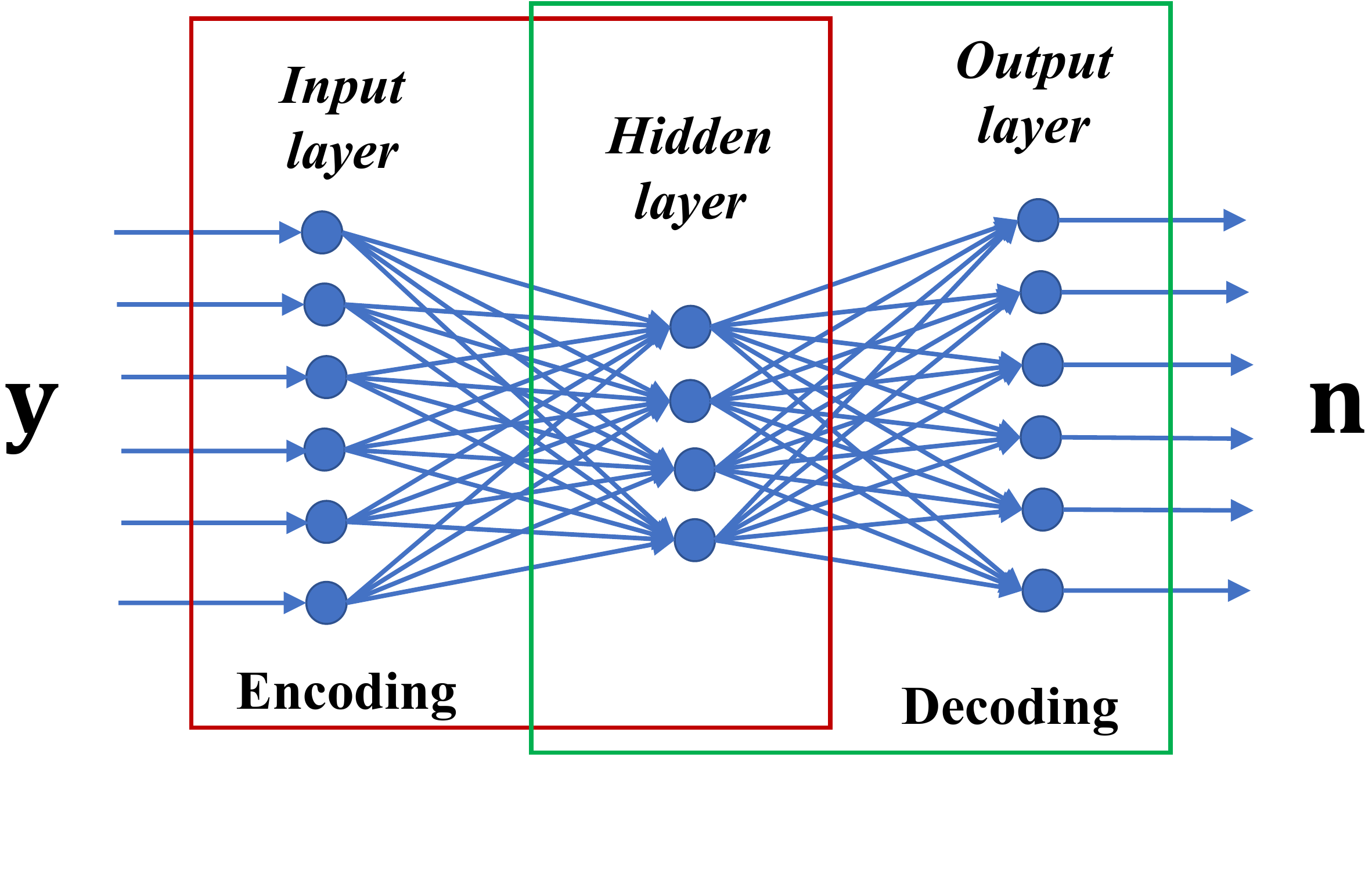}
         \caption{Training phase}
         \label{fig:NE}
     \end{subfigure}
     ~~~~~~~ %\hfill
     \begin{subfigure}[b]{0.34\textwidth} %48
         \centering
         \includegraphics[width=\textwidth]{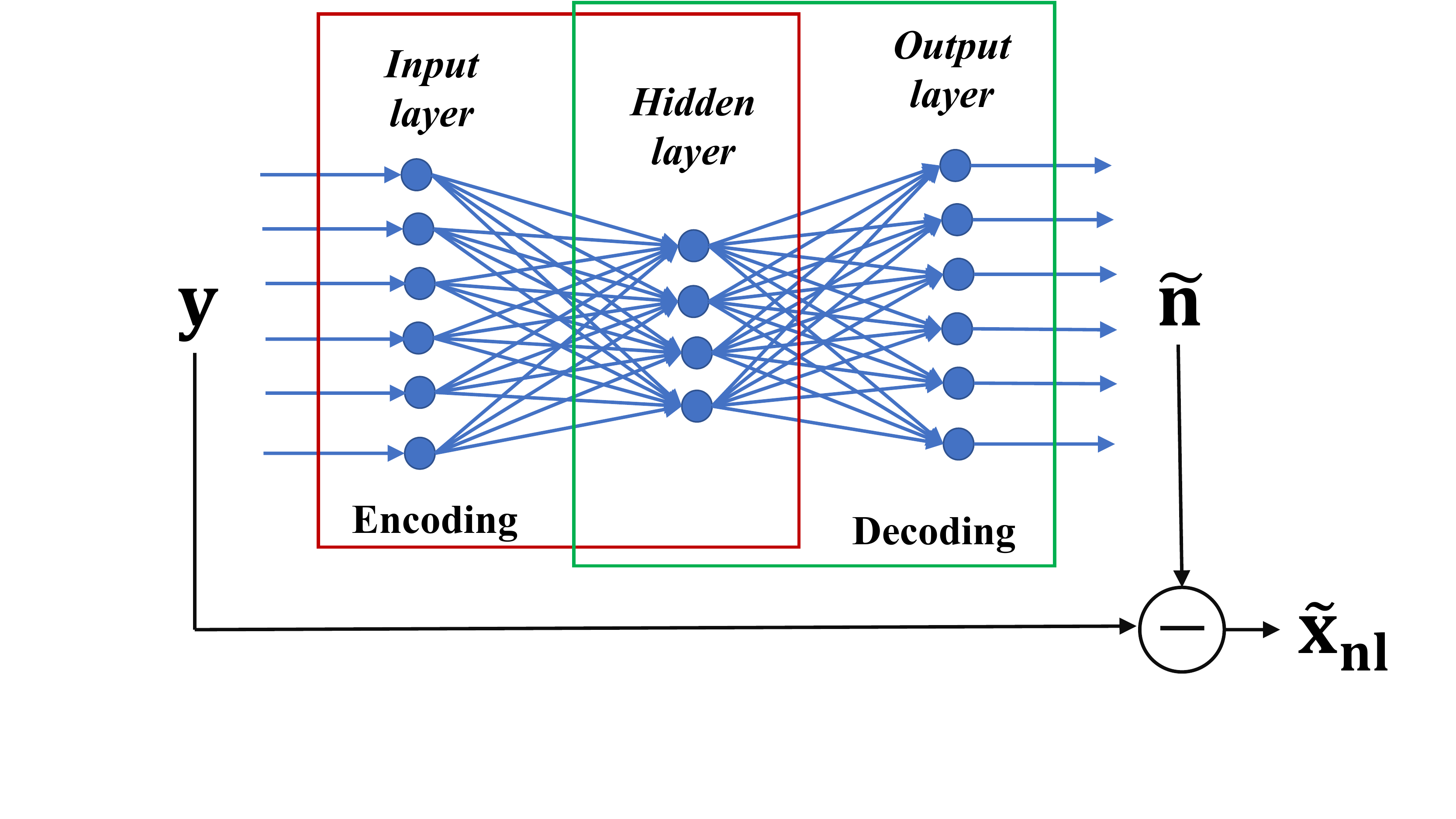}
         \caption{Test phase}
         \label{fig:NP}
     \end{subfigure}
        \caption{An illustration of the concept of nlDAE.}
        \label{nlDAEdefinition}
          %\vspace{-1.5em}

\end{figure*}

%\begin{figure*}
%     \centering
%     \begin{subfigure}[b]{0.48\textwidth}
%         \centering
%         \includegraphics[width=\textwidth]{Fig_conceptA.eps}
%         \caption{{Example 1 and 2: according to $\sigma_N$ (fixed $\mu_N$).}}
%         \label{FigConceptA}
%     \end{subfigure}
%     \hfill
%     \begin{subfigure}[b]{0.48\textwidth}
%         \centering
%         \includegraphics[width=\textwidth]{Fig_conceptB.eps}
%         \caption{{Example 3 and 4: according to $\mu_N$ (fixed $\sigma_N$).}}
%         \label{FigConceptB}
%     \end{subfigure}
%      \caption{{Simple examples of comparison between DAE and nlDAE.}}
%        \label{FigConcept}
%          \vspace{-1.5em}
%\end{figure*}

\begin{figure}[t!]
\centering
\includegraphics[width=0.48\textwidth]{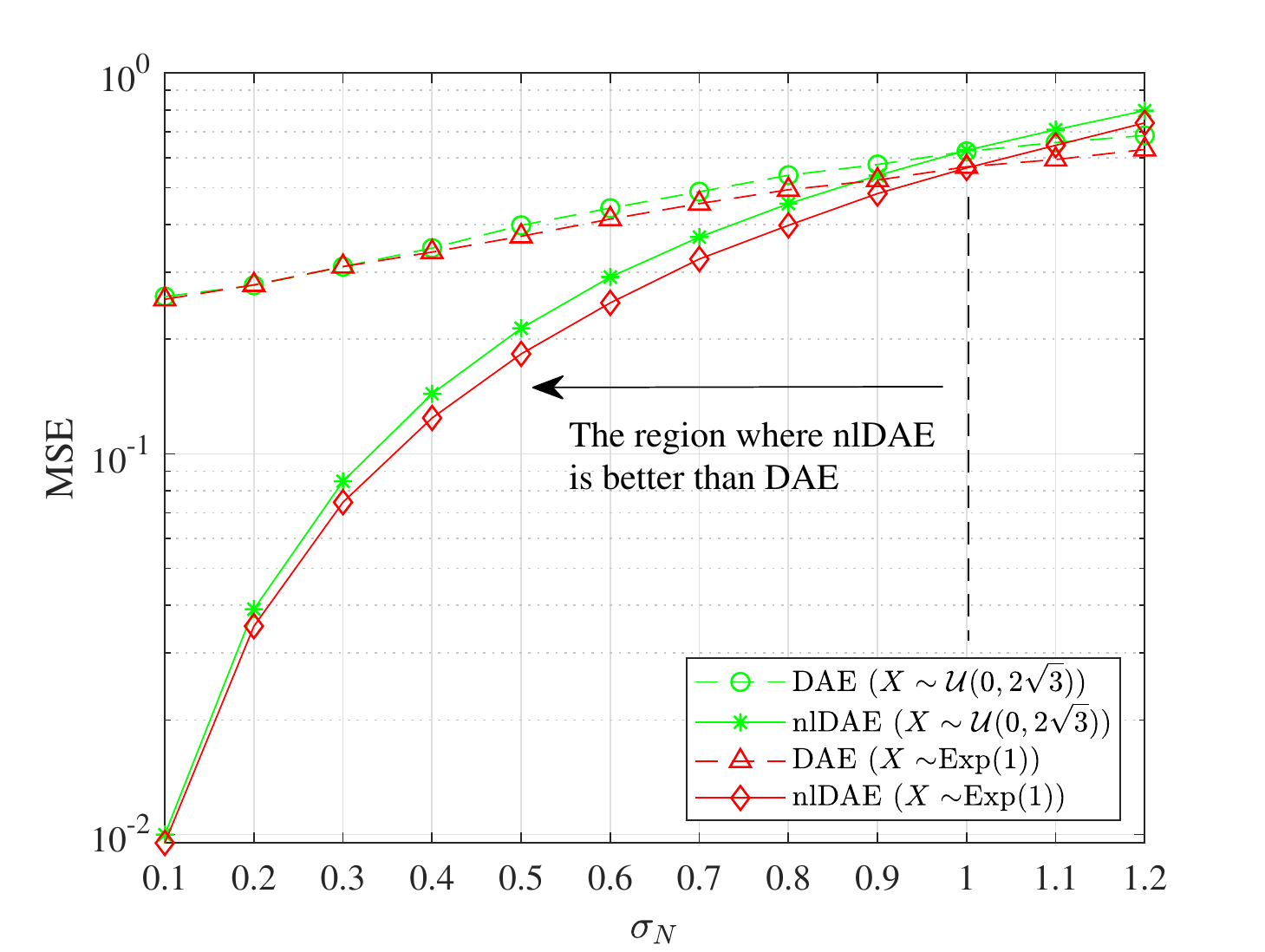}
\caption{A simple example of comparison between DAE and nlDAE: reconstruction error according to $\sigma_N$.}
\label{FigConcept}
\end{figure}

In the traditional estimation problem of signal processing, $N$ is treated as an obstacle to the reconstruction of $X$. Therefore, most of the studies have focused on restoring $X$ as much as possible, which can be expressed as a function of $X$ and $N$. Along with this philosophy, ML-based denoising techniques, e.g., DAE, have also been developed in various signal processing fields with the aim of maximizing the ability to restore $X$ from $Y$. Unlike the conventional approaches, we hypothesize that, if $N$ has a simpler statistical characteristic than $X$, it will be better to subtract from $Y$ after restoring $N$.

We first look into the mechanism of DAE to build neural networks.
Recall that DAE attempts to regenerate the original data $\mathbf{x}$ from the noisy observation $\mathbf{y}$ via training the neural network.
Thus, the parameters of a DAE model can be optimized by minimizing the average reconstruction error in the training phase as follows:
\begin{equation}\label{DAEoptimization}
\theta^*, \theta^{'*} = \argmin_{\theta, \theta^{'}} \frac{1}{M} \sum_{i=1}^{M} \mathcal{L} \big( \mathbf{x}^{(i)} , g_{\theta^{'}}(f_\theta (\mathbf{y}^{(i)})) \big),
\end{equation}
where $\mathcal{L}$ is a loss function such as squared error between two inputs.
Then, the $j$-th regenerated data $\Tilde{\mathbf{x}}^{(j)}$ from ${\mathbf{y}}^{(j)}$ in the test phase can be obtained as follows for all $j \in \{1, \cdots , L\}$:
\begin{equation}\label{regenatingx}
    \Tilde{\mathbf{x}}^{(j)} = g_{\theta^{'*}}(f_{\theta^*} (\mathbf{y}^{(j)})).
\end{equation}

It is noteworthy that, if there are two different neural networks which attempt to regenerate the original data and the noise from the noisy input, the linear summation of these two regenerated data would be different from the input. This means that either $\mathbf{x}$ or $\mathbf{n}$ is more effectively regenerated from $\mathbf{y}$. Therefore, we can hypothesize that learning $N$, instead of $X$, from $Y$ can be beneficial in some cases even if the objective is still to reconstruct $X$. This constitutes the fundamental idea of nlDAE.

The training and test phases of nlDAE are depicted in Fig. \ref{nlDAEdefinition}.
The parameters of nlDAE model can be optimized as follows for all $i \in \{1, \cdots , M\}$:
\begin{equation}
\theta_{nl}^*, {\theta}_{nl}^{'*} = \argmin_{\theta, \theta^{'}} \frac{1}{M} \sum_{i=1}^{M} \mathcal{L} \big( \mathbf{n}^{(i)}, g_{\theta'}(f_{\theta} (\mathbf{y}^{(i)})) \big).
\end{equation}
Notice that the only difference from \eqref{DAEoptimization} is that $\mathbf{x}^{(i)}$ is replaced by $\mathbf{n}^{(i)}$. Let $\Tilde{\mathbf{x}}_{nl}^{(j)}$ denote the $j$-th regenerated data based on nlDAE, which can be represented as follows for all $j \in \{1, \cdots , L\}$:
\begin{equation}
    \Tilde{\mathbf{x}}_{nl}^{(j)} =   \mathbf{y}^{(j)} - g_{\theta_{nl}^{'*}}(f_{\theta_{nl}^*} (\mathbf{y}^{(j)})).
\end{equation}

To provide the readers with insights into nlDAE, we examine two simple examples where the standard deviation of $X$ is fixed as 1, i.e., $\sigma_X=1$, and that of $N$ varies. $Y = X + N$ is comprised as follows:
\begin{itemize}
    \item Example 1: $X\sim \mathcal{U}(0, 2\sqrt{3})$ and $N \sim \mathcal{N}(0, \sigma_N)$.
    \item Example 2: $X\sim \text{Exp}(1)$ and $N \sim \mathcal{N}(0, \sigma_N)$.
\end{itemize}
Fig. \ref{FigConcept} describes the performance comparison between DAE and nlDAE in terms of mean squared error (MSE) for the two examples\footnote{Throughout this letter, the squared error and the scaled conjugate gradient are applied as the loss function and the optimization method, respectively.}.
Here, we set $P=12$, $P'=9$, $M=10000$, and $L=5000$.
It is observed that nlDAE is superior to DAE when $\sigma_N$ is smaller than $\sigma_X$ in Fig. \ref{FigConcept}.
The gap between nlDAE and DAE widens with lower $\sigma_X$. This implies that the standard deviation is an important factor when we select the denoiser between DAE and nlDAE.

These examples show the consideration of whether $X$ or $N$ is easier to be regenerated, which is highly related to differential entropy of each random variable, $H(X)$ and $H(N)$ \cite{marsh2013introduction}. The differential entropy is normally an increasing function over the standard deviation of the corresponding random variable, e.g., $H(N) = \log(\sigma_N \sqrt{2 \pi e})$. Naturally, it is efficient to reconstruct a random variable with a small amount of information, and the standard deviation can be a good indicator.

\section{Case Studies}

To validate the advantage of nlDAE over the conventional DAE in practical problems, we provide three applications for IoT devices in the following subsections.
We assume that the noise follows Bernoulli and normal distributions, respectively, in the first two cases, which are the most common noise modeling.
The third case deals with noise that follows a distribution expressed as a mixture of various random variables.
For all the studied use cases, we select the DAE as the conventional denoiser as a baseline for performance comparison.
We present the case studies in the first three subsections. Then, we discuss the experimental results in Sec. \ref{analysis}.

\subsection{Case Study I: Signal Restoration}
In this use case, the objective is to recover the original signal from the noisy signal which is modeled by the corruptions over samples.

\subsubsection{Model}
The sampled signal of randomly superposed sinusoids, e.g., the recorded acoustic wave, is the summation of samples of $k$ damped sinusoidal waves which can be represented as follows:
\begin{equation}
    \mathbf{x}=\Big\{\sum_{l=1}^{k}V_l e^{-\gamma_l n\Delta t} \cos(2\pi f_l n\Delta t) \Big\}_{n=0}^{P-1},
\end{equation}
where $V_l$, $\gamma_l$, and $f_l$ are the peak amplitude, the damping factor, and the frequency of the $l$-th signal, respectively.
Here, the time interval for sampling, $\Delta t$, is set to satisfy the Nyquist theorem, i.e., $\frac{1}{2\Delta t} > \max \{ f_1, \cdots , f_k \}$.
To consider the corruption of $\mathbf{x}$, let us assume that the probability of corruption for each sample follows the Bernoulli distribution $\text{Ber}(p_{cor})$, which indicates the corruption with the probability $p_{cor}$. In addition, let $\mathbf{b} \in \{0,1\}^P$ denote  the realization of $\text{Ber}(p_{cor})$ over $P$ samples. Naturally, the corrupted signal, $\mathbf{y} \in \mathbb{R}^P$, can be represented as follows:

\begin{equation}\label{csb}
\mathbf{y}=\mathbf{x} + C\mathbf{b},
\end{equation}
where $C$ is a constant representing the sample corruption.

\subsubsection{Application of nlDAE}
Based on \eqref{csb}, the denoised signal $\Tilde{\mathbf{x}}_{nl}^{(j)}$ can be represented by
\begin{equation}
    \Tilde{\mathbf{x}}_{nl}^{(j)} =   \mathbf{x}^{(j)} + C\mathbf{b}^{(j)} - g_{\theta_{nl}^{'*}}(f_{\theta_{nl}^*} (\mathbf{x}^{(j)} + C\mathbf{b}^{(j)}),
\end{equation}
where
\begin{equation*}
\theta_{nl}^*, {\theta}_{nl}^{'*} = \argmin_{\theta, \theta^{'}} \frac{1}{M} \sum_{i=1}^{M} \mathcal{L} \big( C\mathbf{b}^{(i)}, g_{\theta'}(f_{\theta} (\mathbf{x}^{(i)} + C\mathbf{b}^{(i)})) \big).
\end{equation*}
\subsubsection{Experimental Parameters}
We evaluate the performance of the proposed nlDAE in terms of the MSE of restoration.
For the experiment, the magnitude of noise $C$ is set to 1 for simplicity. In addition, $V_l$, $\gamma_l$, and $f_l$ follow $\mathcal{N}(0,1)$, $\mathcal{U}(0,10^{3})$, and $\mathcal{U}(0,10 \text{ kHz})$, respectively, for all $l$. The sampling time interval $\Delta t$ is set to $0.5 \times 10^{-4}$ second, and the number of samples $P$ is $12$. We set $P' = 9$, $p_{cor}=0.9$, and $M = 10000$ unless otherwise specified.

\subsection{Case Study II: Symbol Demodulation}

Here, the objective is to improve the symbol demodulation quality through denoising the received signal that consists of channel, symbols, and additive noise.

\subsubsection{Model}

Consider an orthogonal frequency-division multiplexing (OFDM) system with $P$ subcarriers where the subcarrier spacing is expressed by $\Delta f$.
Let $\mathbf{d} \in \mathbb{C}^P$ be a sequence in frequency domain. $\mathbf{d}[n]$ is the $n$-th element of $\mathbf{d}$ and denotes the symbol transmitted over the $n$-th subcarrier.
In addition, let $K$ denote the pilot spacing for channel estimation.
%For brevity, we neglect the cyclic-prefix, i.e.,, the symbol duration is ${1}/{\Delta f}$.
Furthermore, the channel impulse response (CIR) can be modeled by the sum of Dirac-delta functions as follows:
\begin{equation}
    h(t,\tau)=\sum_{l=0}^{L_p-1} \alpha_l \delta(t-\tau_l),
\end{equation}
where $\alpha_l$, $\tau_l$, and $L_p$ are the complex channel gain, the excess delay of $l$-th path, and the number of multipaths, respectively.
Let $\mathbf{x} \in \mathbb{C}^P$ denote the discrete signal obtained by $P$-point fast Fourier transform (FFT) after the sampling of the signal experiencing the channel at the receiver, which can be represented as follows:
\begin{equation}{\label{ds}}
    \mathbf{x} = \mathbf{d} \odot \mathbf{h} = \{ \mathbf{d}[n]\sum_{l=0}^{L_p-1} \alpha_l e^{-j2\pi n \Delta f \tau_l}\}_{n=0}^{P-1},
\end{equation}
where $\odot$ denotes the operator of the Hadamard product. Here, $\mathbf{h} \in \mathbb{C}^P$ is the channel frequency response (CFR), which is the $P$-point FFT of $h(t,\tau)$.
In addition, let $\mathbf{n} \in \mathbb{C}^P$ denote the realization of the random variable $N \sim \mathcal{CN}(0,\sigma_N)$. Finally, $\mathbf{y}(=\mathbf{d}\odot \mathbf{h} + \mathbf{n})$ is the noisy observed signal.

Our goal is to minimize the symbol error rate (SER) over $\mathbf{d}$ by maximizing the quality of denoising $\mathbf{y}$. We assume the method of channel estimation is fixed as the cubic interpolation \cite{agarwal2012pilot} to focus on the performance of denoising the received signal.

\begin{figure*}
     \centering
     \begin{subfigure}[b]{0.24\textwidth}
         \centering
         \includegraphics*[trim=0.17cm 0.1cm 0.8cm 0.5cm, clip=true, width=\textwidth]{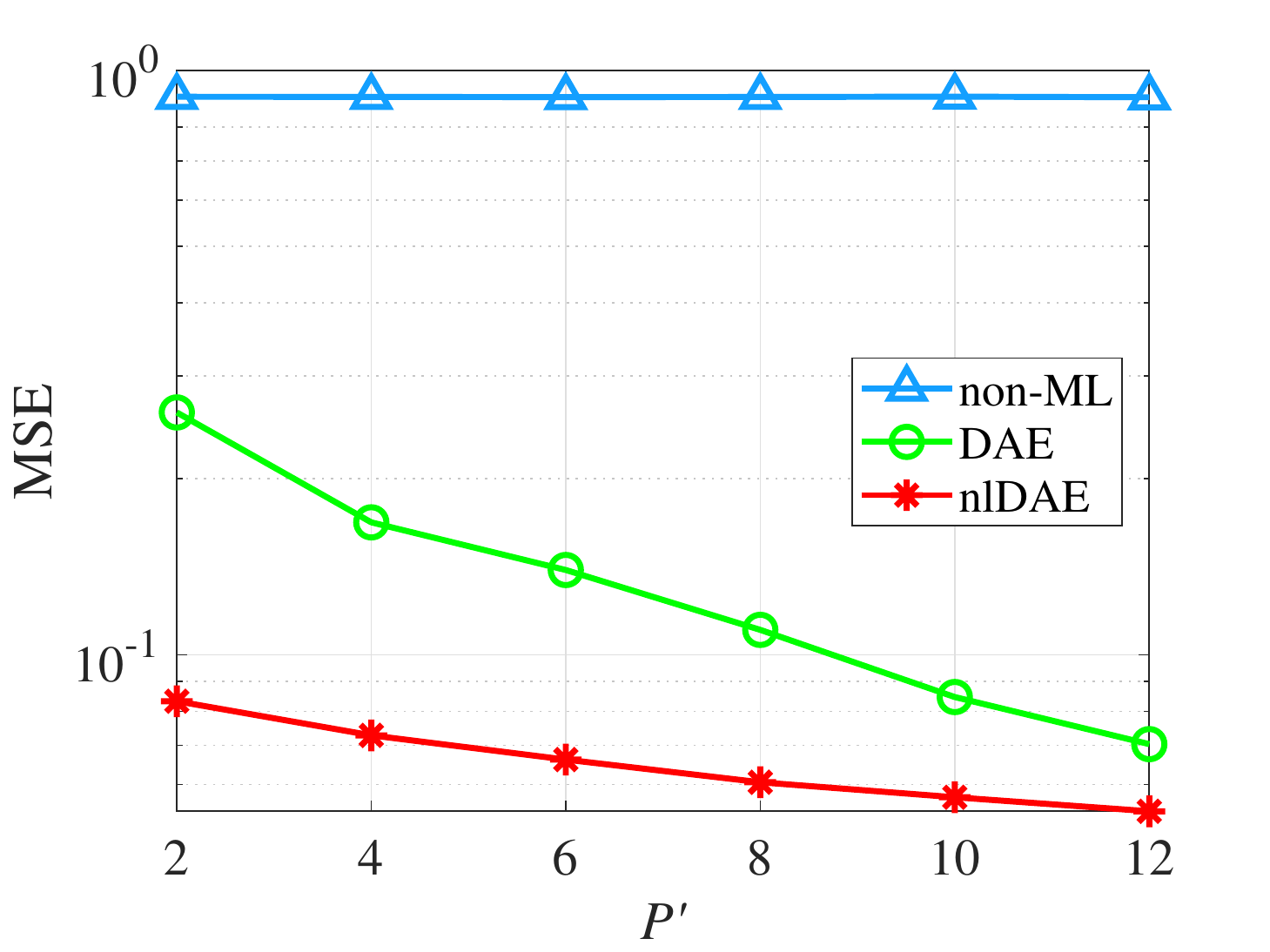}
         \caption{}
         \label{fig:NE_mse_P}
     \end{subfigure}
     \hfill
     \begin{subfigure}[b]{0.24\textwidth}
         \centering
         \includegraphics*[trim=0.17cm 0.1cm 0.8cm 0.5cm, clip=true, width=\textwidth]{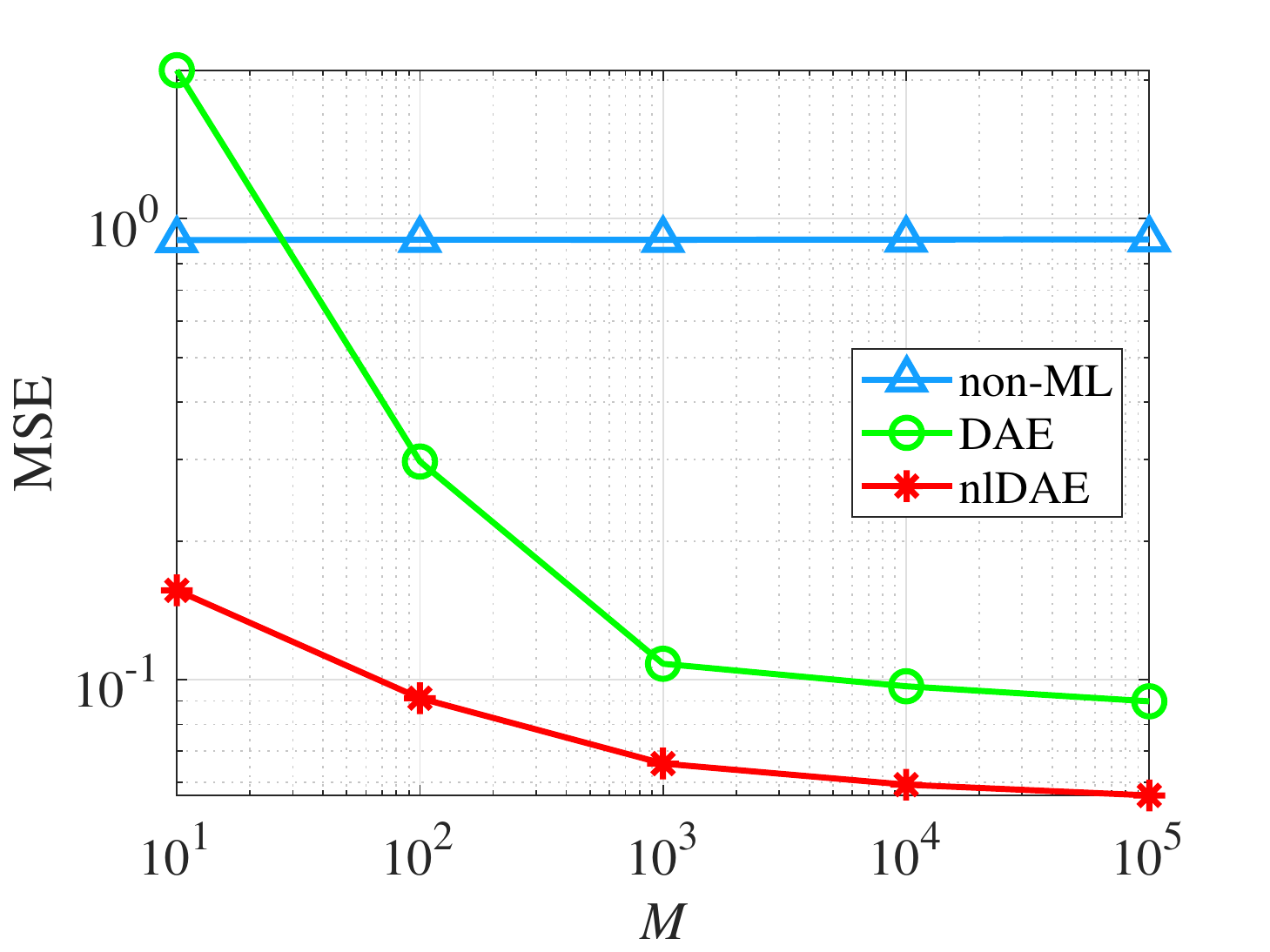}
         \caption{}
         \label{fig:NP_mse_M}
     \end{subfigure}
     ~%\\
     \begin{subfigure}[b]{0.24\textwidth}
         \centering
         \includegraphics*[trim=0.17cm 0.1cm 0.8cm 0.5cm, clip=true, width=\textwidth]{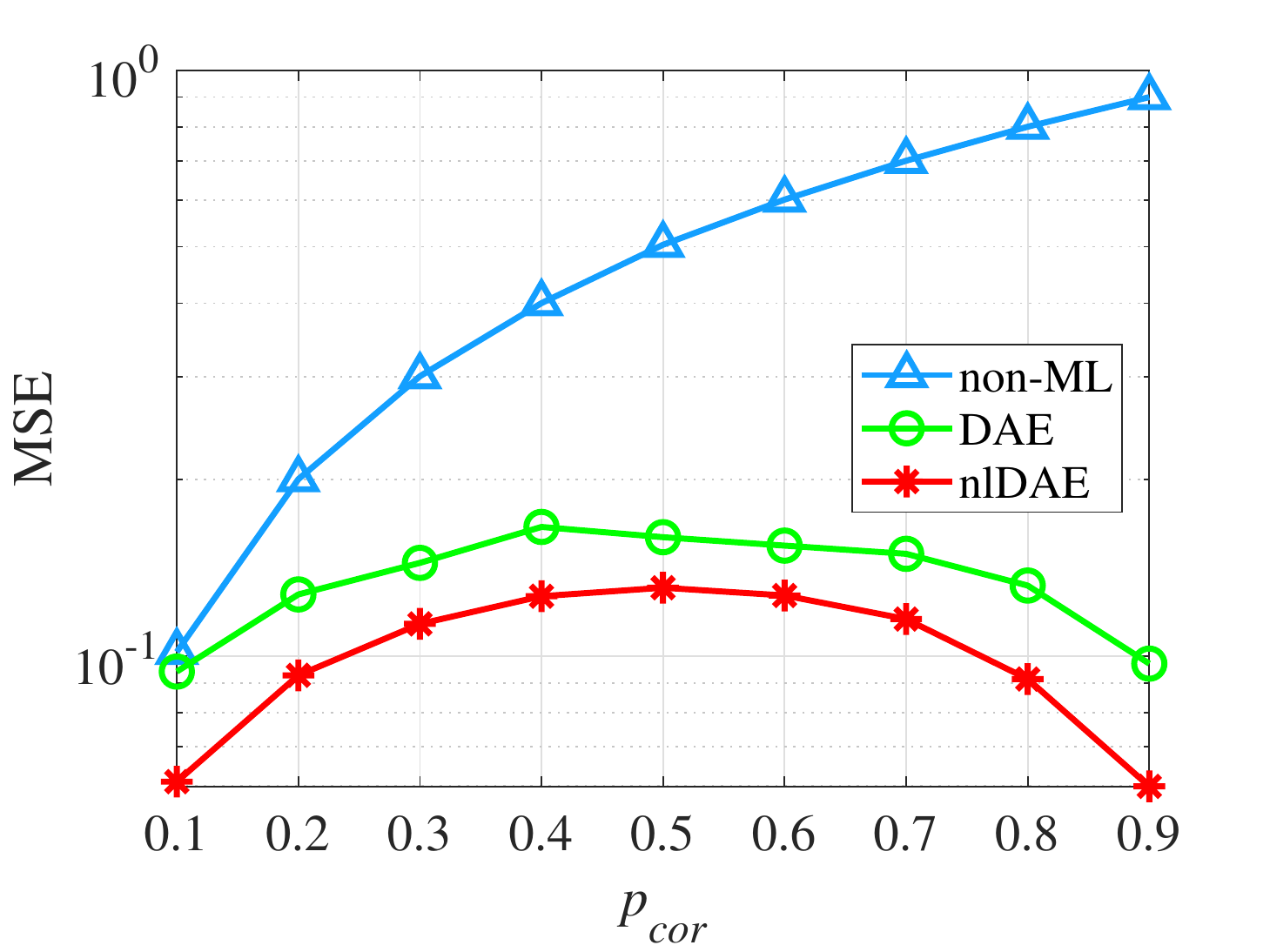}
         \caption{}
         \label{fig:NP_mse_SNR}
     \end{subfigure}
     \hfill
     \begin{subfigure}[b]{0.24\textwidth}
         \centering
         \includegraphics*[trim=0.17cm 0.1cm 0.6cm 0.5cm, clip=true, width=\textwidth]{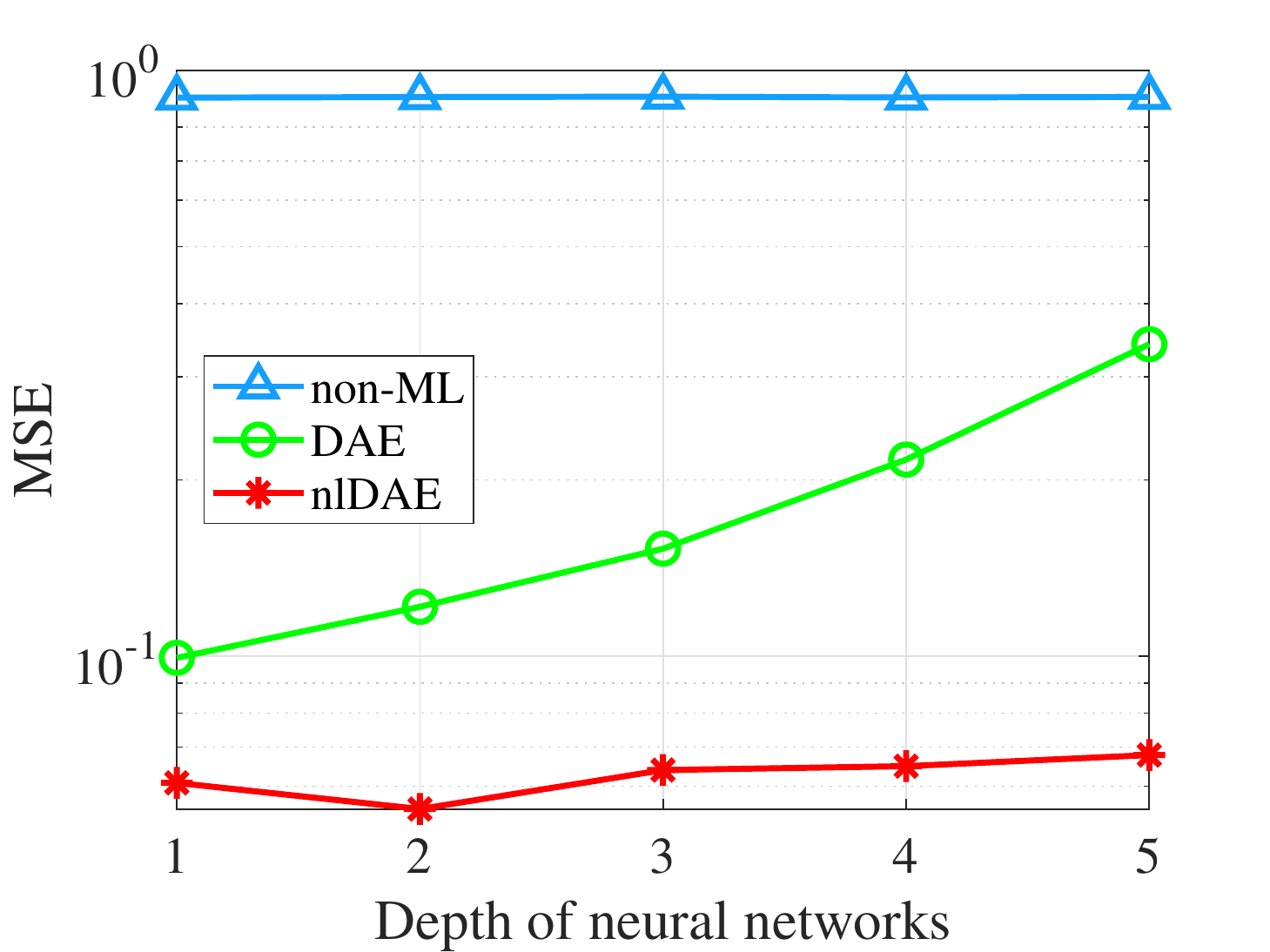}
         \caption{}
         \label{fig:NP_mse_Depth}
     \end{subfigure}
             \caption{Case study I (signal restoration): MSE according to (a) the dimension of latent space; (b) the size of training dataset; (c) $p_{cor}$; and (d) the depth of neural networks.}
        \label{fig:three graphs_mse}
\end{figure*}

\begin{figure*}
     \centering
     \begin{subfigure}[b]{0.24\textwidth}
         \centering
         \includegraphics*[trim=0.17cm 0.1cm 0.8cm 0.5cm, clip=true, width=\textwidth]{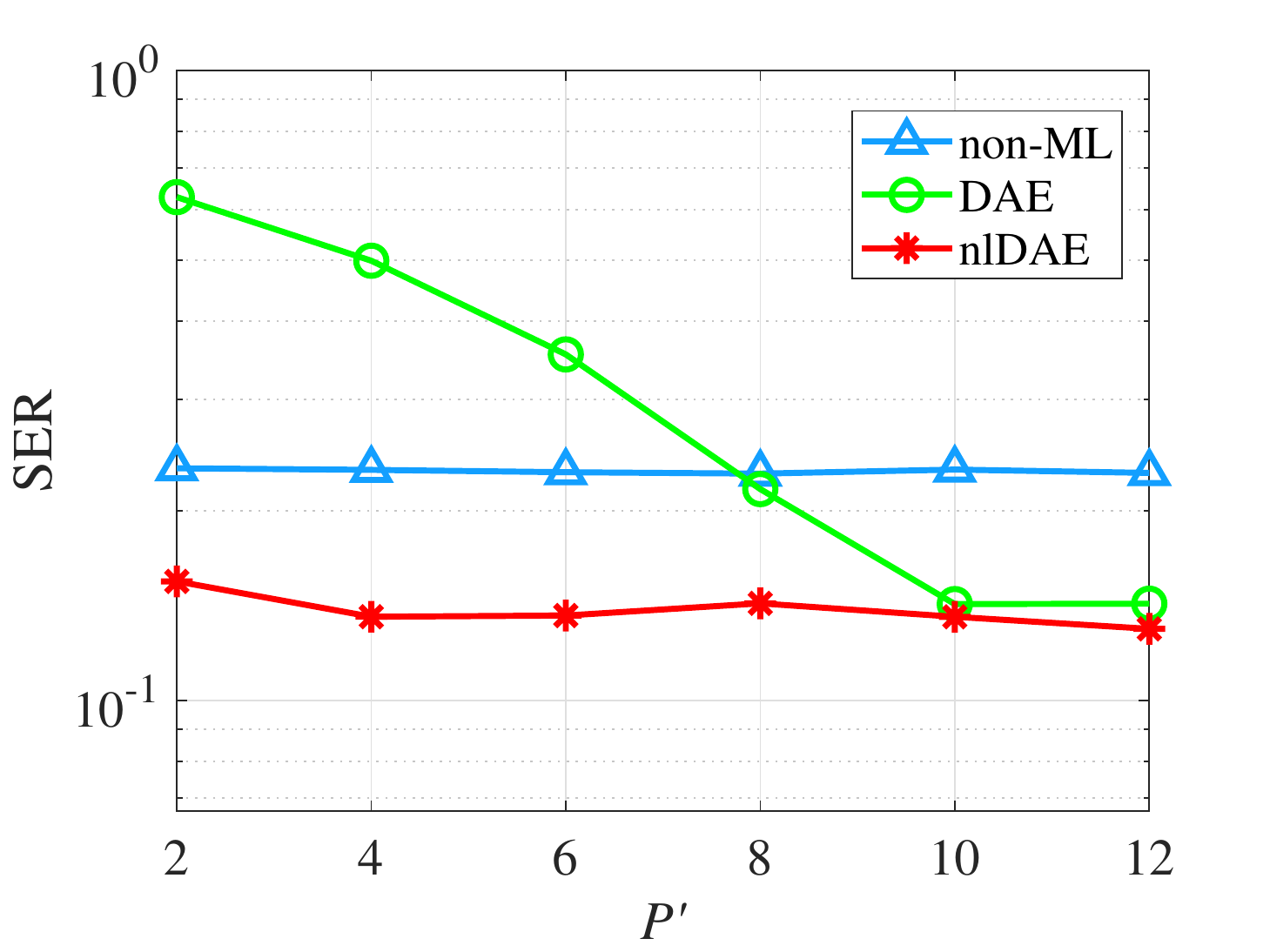}
         \caption{}
         \label{fig:NE_ser_P}
     \end{subfigure}
     \hfill
     \begin{subfigure}[b]{0.24\textwidth}
         \centering
         \includegraphics*[trim=0.17cm 0.1cm 0.8cm 0.5cm, clip=true, width=\textwidth]{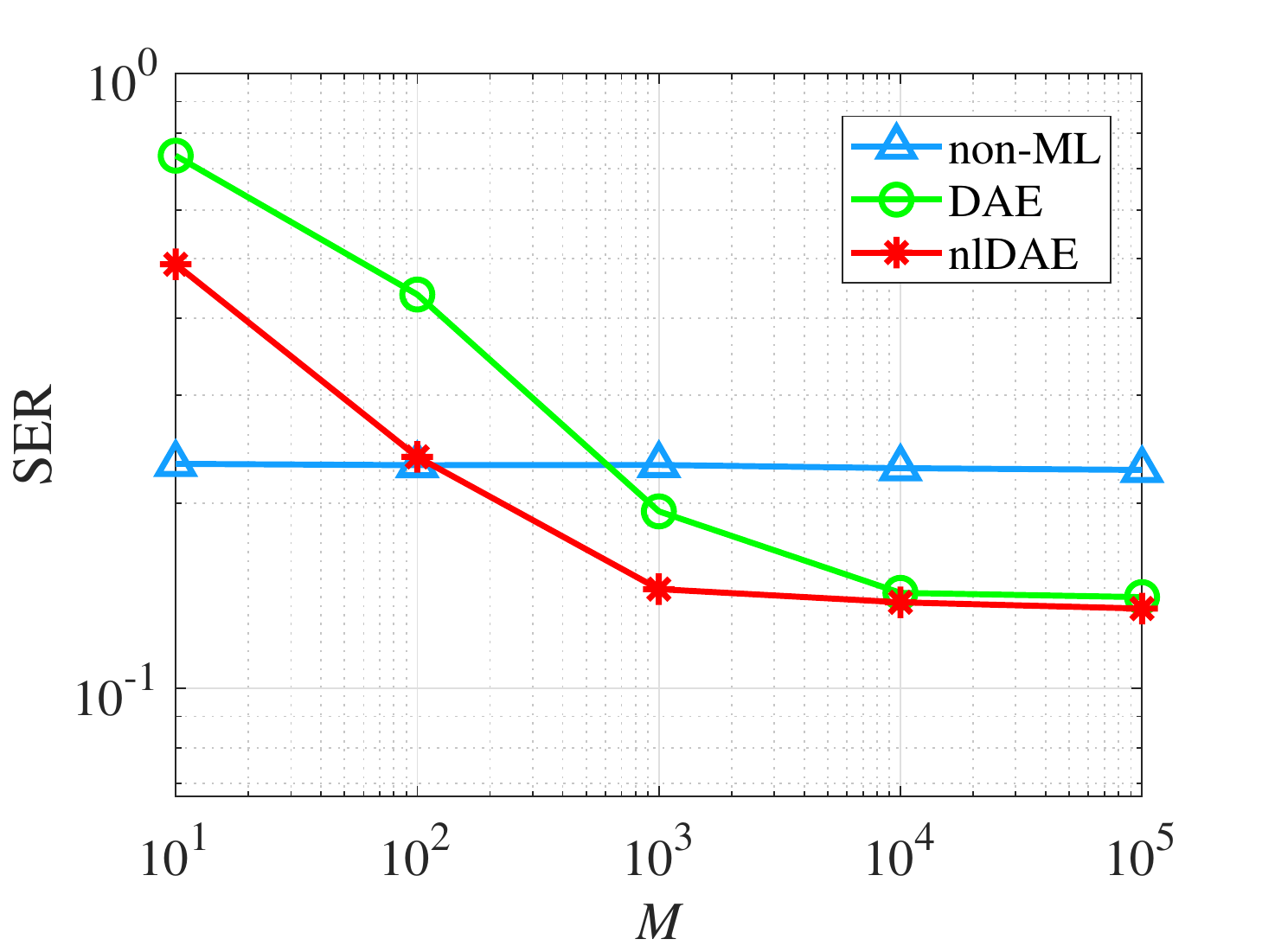}
         \caption{}
         \label{fig:NP_ser_M}
     \end{subfigure}
     ~%\\
     \begin{subfigure}[b]{0.24\textwidth}
         \centering
         \includegraphics*[trim=0.17cm 0.1cm 0.8cm 0.5cm, clip=true, width=\textwidth]{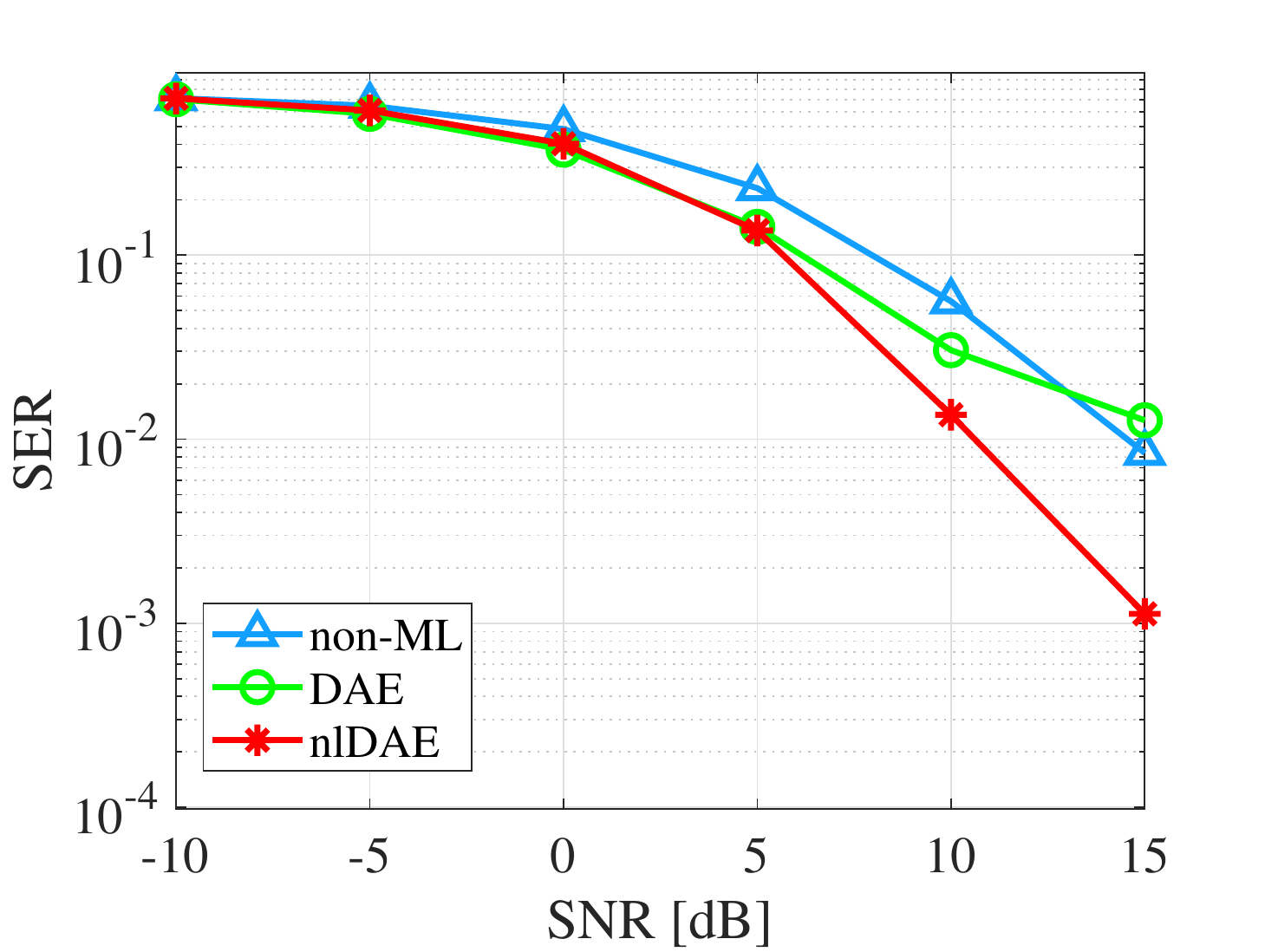}
         \caption{}
         \label{fig:NP_ser_SNR}
     \end{subfigure}
     \hfill
     \begin{subfigure}[b]{0.24\textwidth}
         \centering
         \includegraphics*[trim=0.17cm 0.1cm 0.8cm 0.5cm, clip=true, width=\textwidth]{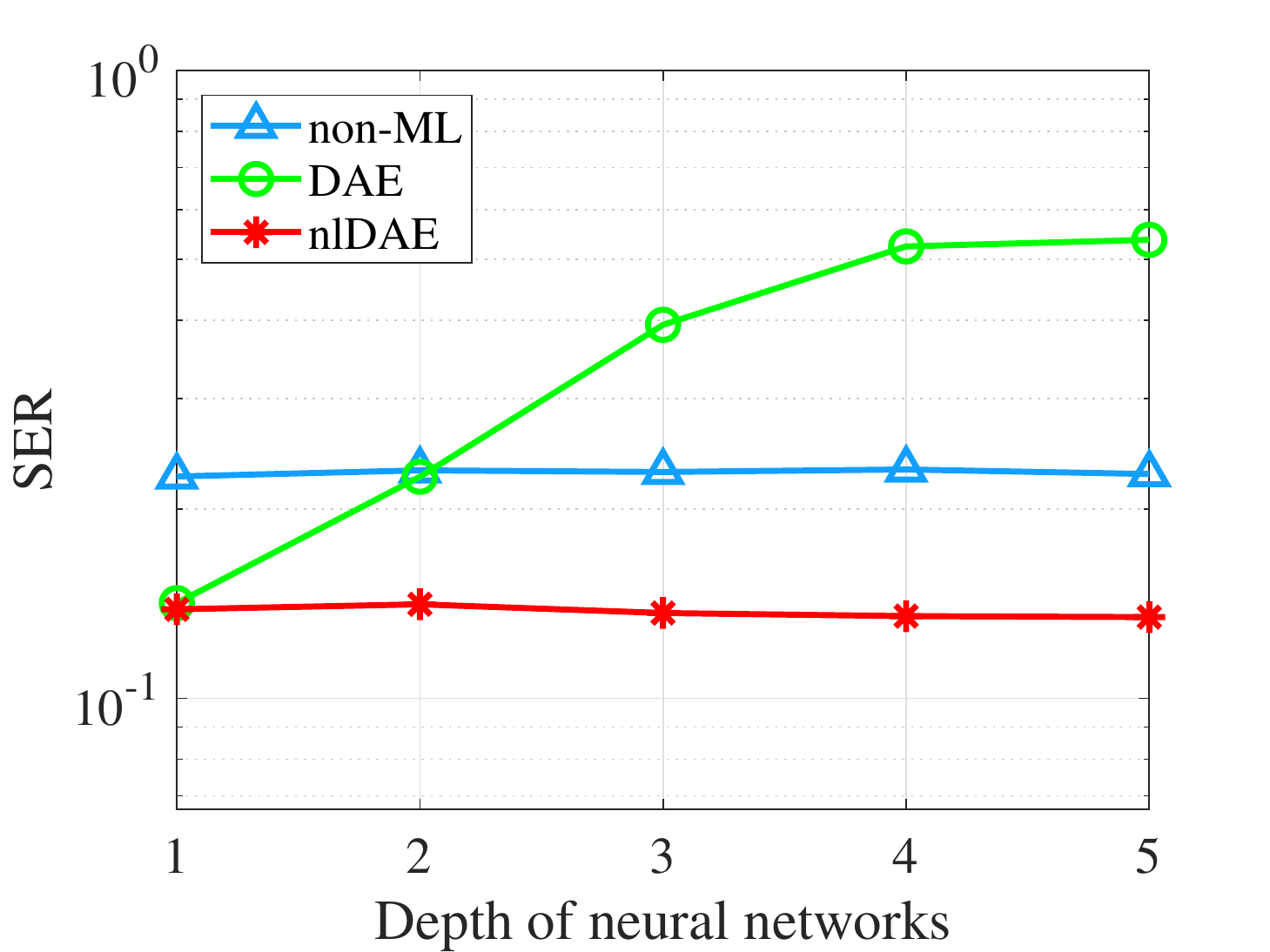}
         \caption{}
         \label{fig:NP_ser_Depth}
     \end{subfigure}
             \caption{Case study II (symbol demodulation): SER according to (a) the dimension of latent space; (b) the size of training dataset; (c) SNR; and (d) the depth of neural networks.}
        \label{fig:three graphs_ser}
\end{figure*}
\begin{figure*}
     \centering
     \begin{subfigure}[b]{0.24\textwidth}
         \centering
         \includegraphics*[trim=0.17cm 0.1cm 0.8cm 0.5cm, clip=true, width=\textwidth]{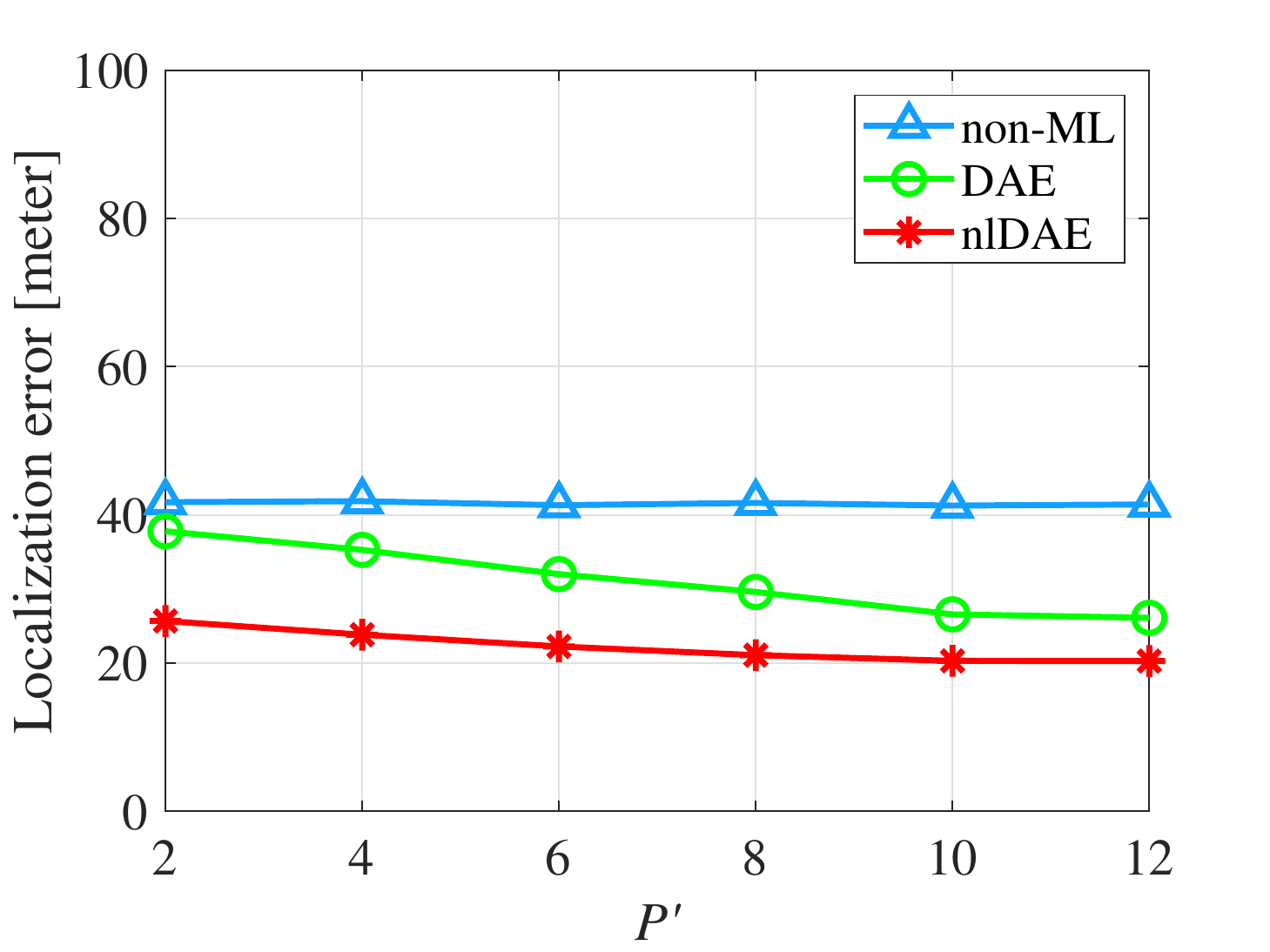}
         \caption{}
         \label{fig:NE_loc_P}
     \end{subfigure}
    \hfill
     \begin{subfigure}[b]{0.24\textwidth}
         \centering
         \includegraphics*[trim=0.17cm 0.1cm 0.8cm 0.5cm, clip=true, width=\textwidth]{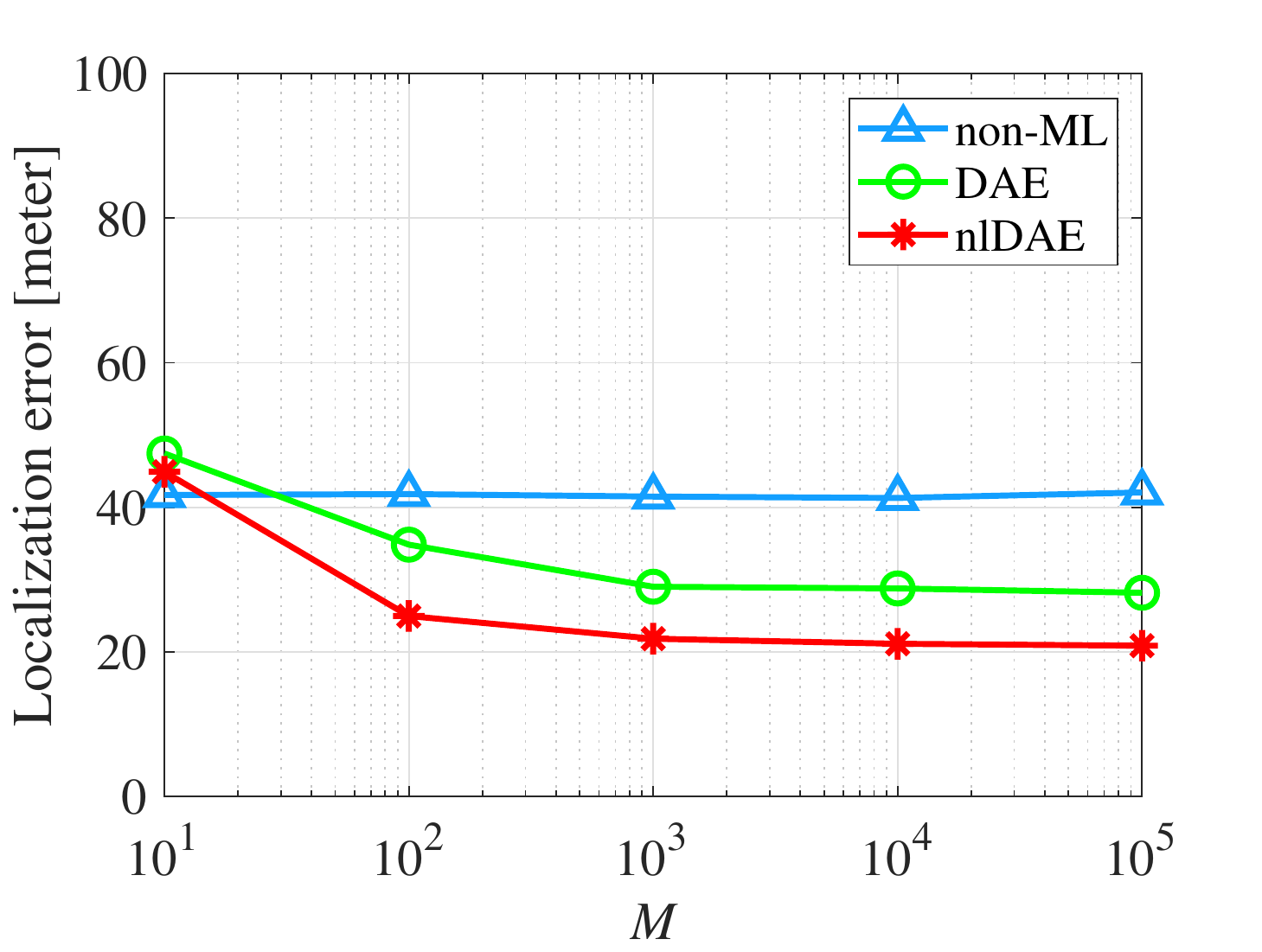}
         \caption{}
         \label{fig:NP_loc_M}
     \end{subfigure}
     ~%\\
     \begin{subfigure}[b]{0.24\textwidth}
         \centering
         \includegraphics*[trim=0.17cm 0.1cm 0.8cm 0.5cm, clip=true, width=\textwidth]{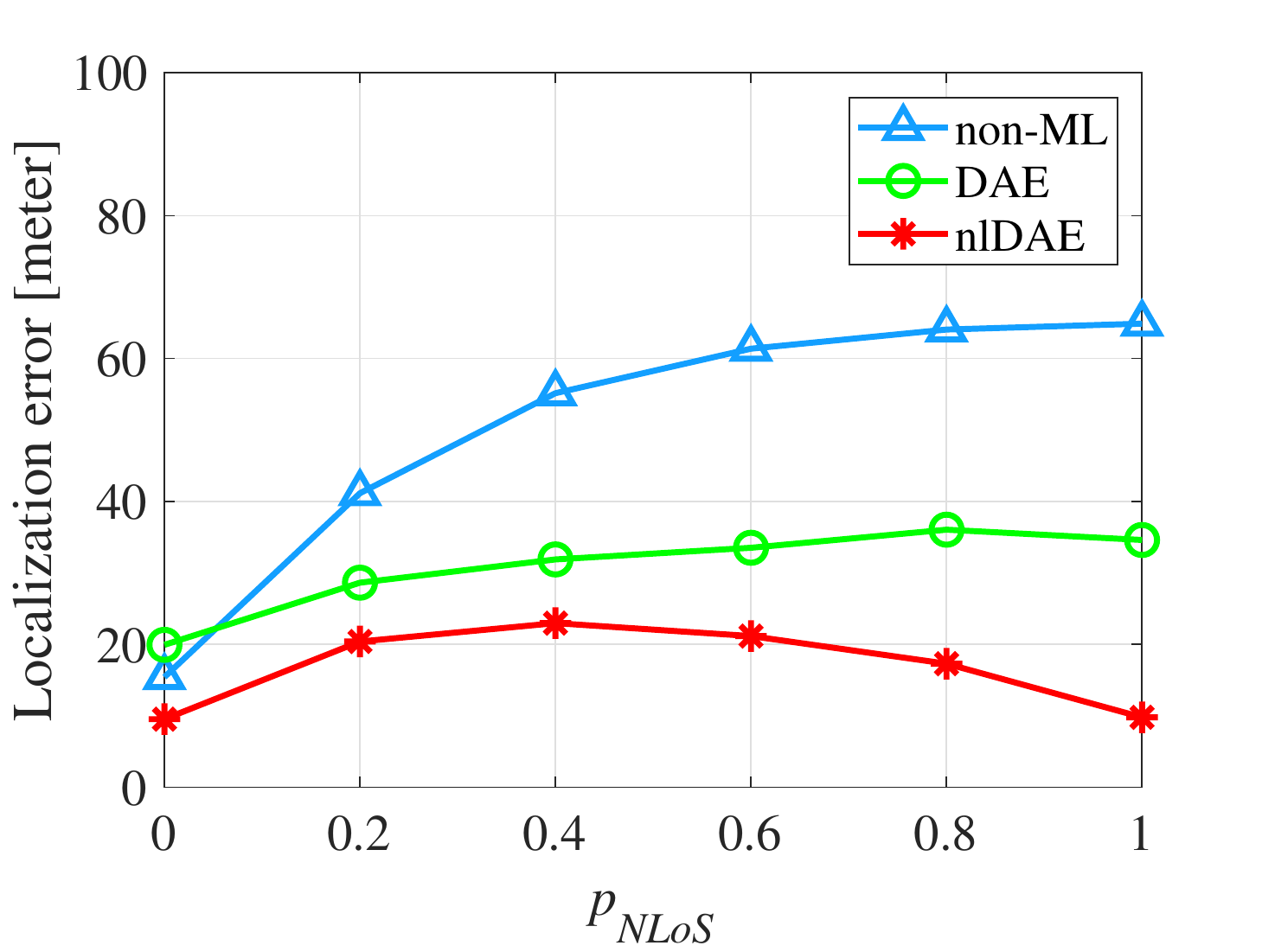}
         \caption{}
         \label{fig:NP_loc_prob}
     \end{subfigure}
     \hfill
     \begin{subfigure}[b]{0.24\textwidth}
         \centering
         \includegraphics*[trim=0.17cm 0.1cm 0.8cm 0.5cm, clip=true, width=\textwidth]{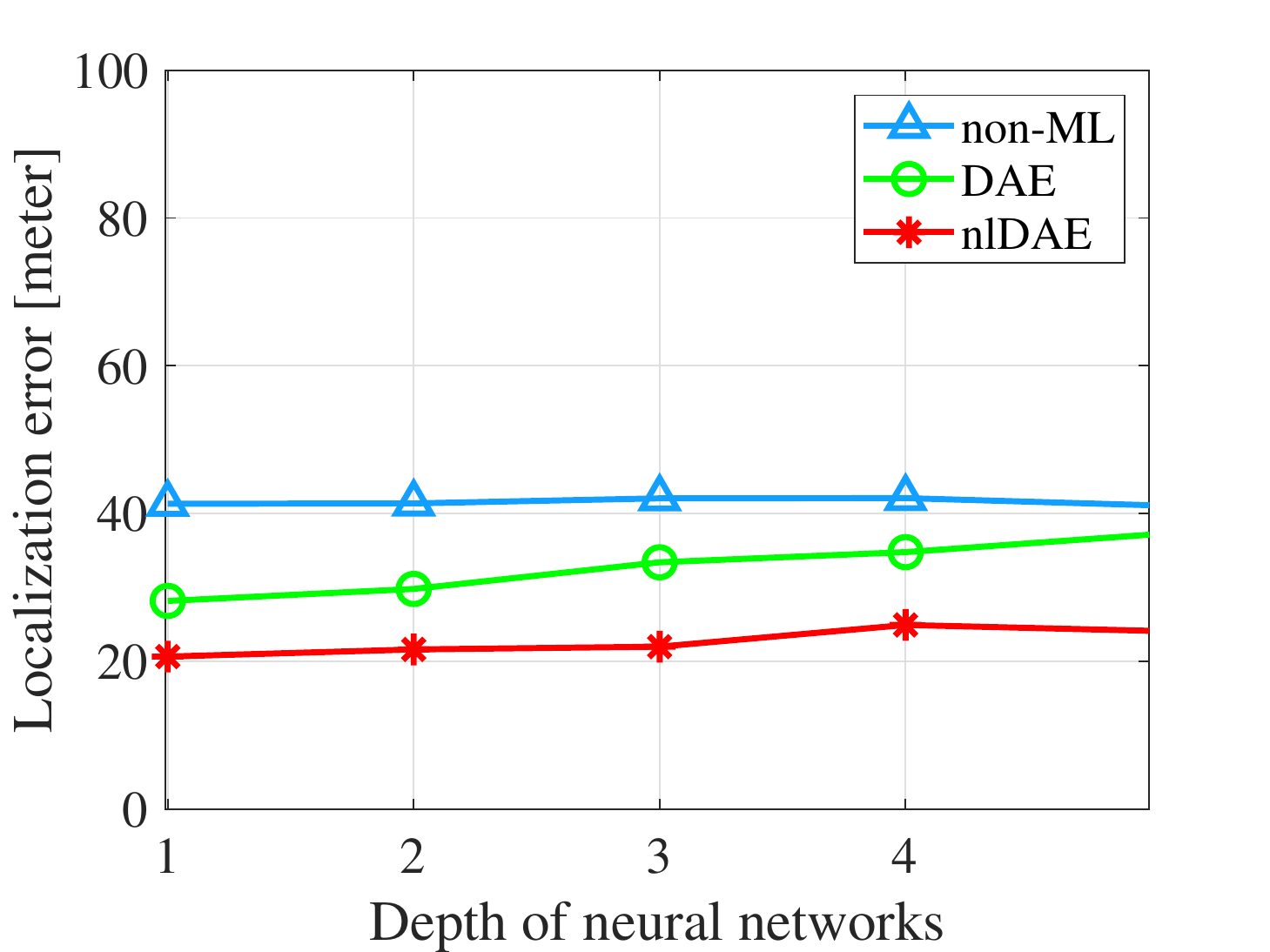}
         \caption{}
         \label{fig:NP_loc_Depth}
     \end{subfigure}
             \caption{Case study III (precise localization): Localization error according to (a) the dimension of latent space; (b) the size of training dataset; (c) $p_{NLoS}$; and (d) the depth of neural networks.}
        \label{fig:three graphs_loc}
\end{figure*}

\subsubsection{Application of nlDAE}

To consider the complex-valued data, we separate it into real and imaginary parts.
$\Re$ and $\Im$ denote the operators capturing real and imaginary parts of an input, respectively.
Thus, $\Tilde{\mathbf{x}}_{nl}^{(j)}$ is the regenerated $\mathbf{d}^{(j)} \odot \mathbf{h}^{(j)}$ by denoising $\mathbf{y}^{(j)}$, which can be represented by
\begin{equation}
\begin{split}
\Tilde{\mathbf{x}}_{nl}^{(j)} & =  \Re (  \mathbf{y}^{(j)}) - g_{\theta_{nl,R}^{'*}}(f_{\theta_{nl,R}^*} ( \Re(\mathbf{y}^{(j)})) ) \\ & + i (\Im (  \mathbf{y}^{(j)}) - g_{\theta_{nl,I}^{'*}}(f_{\theta_{nl,I}^*} ( \Im(\mathbf{y}^{(j)})) )), \end{split}
\end{equation}
where
\begin{equation*}
\begin{split}
\theta_{nl,{R}}^*, {\theta}_{nl,{R}}^{'*} = \argmin_{\theta, \theta^{'}} \frac{1}{M} \sum_{i=1}^{M} \mathcal{L} \big( \Re ( \mathbf{n}^{(i)}), g_{\theta'}(f_{\theta} ( \Re ( \mathbf{y}^{(i)}) ) ) \big), \\
\theta_{nl,{I}}^*, {\theta}_{nl,{I}}^{'*} = \argmin_{\theta, \theta^{'}} \frac{1}{M} \sum_{i=1}^{M} \mathcal{L} \big( \Im ( \mathbf{n}^{(i)}), g_{\theta'}(f_{\theta} ( \Im ( \mathbf{y}^{(i)}) ) ) \big).
\end{split}
\end{equation*}
{Finally, the receiver estimates $\mathbf{h}$ with the predetermined pilot symbols, i.e., $\mathbf{d}[nK+1]$ where $n=0,1,\cdots$, and demodulates $\mathbf{d}$ based on the estimate of $\mathbf{h}$ and the regenerated $\Tilde{\mathbf{x}}_{nl}$.}
\subsubsection{Experimental Parameters}
{The performance of the proposed nlDAE is evaluated when $L=5000$.}
For the simulation parameters, we set $4$ QAM, $P=12$, $\Delta f = 15$ kHz, $L_p = 4$, and $K=3$.
We further assume that $\alpha \sim \mathcal{CN}(0,1)$ and $\tau \sim \mathcal{U}(0,10^{-6})$.
Furthermore, $P' = 9$, SNR$=5$ dB, and $M = 10000$ unless otherwise specified.
We also provide the result of non-ML (i.e., only cubic interpolation).

\subsection{Case Study III: Precise Localization}

The objective of this case study is to improve the localization quality through denoising the measured distance which is represented by the quantized value of the mixture of the true distance and error factors.

\subsubsection{Model} Consider a 2-D localization where $P$ reference nodes and a single target node are randomly distributed.
We estimate the position of the target node with the knowledge of the locations of $P$ reference nodes.
Let $\mathbf{x} \in \mathbb{R}^P$ denote the vector of true distances from $P$ reference nodes to the target node when $X$ denotes the distance between two random points in a 2-D space.
We consider three types of random variables for the noise added to the true distance as follows:
\begin{itemize}
    \item $N_N$: ranging error dependent on signal quality.
    \item $N_U$: ranging error due to clock asynchronization.
    \item $N_B$: non line-of-sight (NLoS) event.
\end{itemize}
We assume that $N_N$, $N_U$, $N_B$ follow the normal, uniform, and Bernoulli distributions, respectively. Hence, we can define the random variable for the noise $N$ as follows:
\begin{equation}
    N = N_N + N_U + R_{NLoS}N_B,
\end{equation}
where $R_{NLoS}$ is the distance bias in the event of NLoS. {Note that $N$ does not follow any known probability distribution because it is a convolution of three different distributions.} Besides, we assume that the distance is measured by time of arrival (ToA). Thus, we define the quantization function $\mathcal{Q}_B$ to represent the measured distance with the resolution of $B$, e.g., $\mathcal{Q}_{10}(23)=20$.
In addition, the localization method based on multi-dimensional scaling (MDS) is utilized to estimate the position of the target node \cite{dokmanic2015euclidean}.

\subsubsection{Application of nlDAE} In this case study, we consider the discrete values quantized by the function $\mathcal{Q}_B$. Here, $\Tilde{\mathbf{x}}_{nl}^{(j)}$ can be represented as follows:
\begin{equation}
\Tilde{\mathbf{x}}_{nl}^{(j)} =  {\mathcal{Q}_B(\mathbf{y}}^{(j)}) - g_{\theta_{nl,R}^{'*}}(f_{\theta_{nl,R}^*} ( {\mathcal{Q}_B(\mathbf{y}}^{(j)}))),
\end{equation}
where
\begin{equation*}
\theta_{nl,{R}}^*, {\theta}_{nl,{R}}^{'*} = \argmin_{\theta, \theta^{'}} \frac{1}{M} \sum_{i=1}^{M} \mathcal{L} \big( \mathcal{Q}_B ( \mathbf{n}^{(i)}), g_{\theta'}(f_{\theta} ( \mathcal{Q}_B ( \mathbf{y}^{(i)}) ) ) \big).
\end{equation*}
Thus, $\Tilde{\mathbf{x}}_{nl}$ is utilized for the estimation of the target node position in nlDAE-assisted MDS-based localization.
\subsubsection{Experimental Parameters} The performance of the proposed nlDAE is evaluated via $L=5000$. In this simulation, $12$ reference nodes and one target node are uniformly distributed in a $100\times 100$ square. We assume that $N_{N} \sim \mathcal{N}(0,10), N_{U} \sim \mathcal{U}(0,20), N_B \sim \text{Ber}(0.2)$, and $R_{NLoS} = 50$. The distance resolution $B$ is set to $10$ for the quantization function $\mathcal{Q}_B$.
Note that $P' = 9$, $p_{NLoS}=0.2$, and $M = 10000$ unless otherwise specified.
We also provide the result of non-ML (i.e., only MDS based localization).

\subsection{Analysis of Experimental Results}\label{analysis}

Fig. \ref{fig:three graphs_mse}(a), Fig. \ref{fig:three graphs_ser}(a), and Fig. \ref{fig:three graphs_loc}(a) show the performance of the three case studies with respect to $P'$, respectively. nlDAE outperforms non-ML and DAE for all ranges of $P'$. Particularly with small values of $P'$, nlDAE continues to perform well, whereas DAE loses its merit. This means that nlDAE provides a good denoising performance even with an extremely small dimension of latent space if the training dataset is sufficient.

The impact of the size of training dataset is depicted in Fig. \ref{fig:three graphs_mse}(b), Fig. \ref{fig:three graphs_ser}(b), and Fig. \ref{fig:three graphs_loc}(b). nlDAE starts to outperform non-ML with $M$ less than 100. Conversely, DAE requires about an order higher $M$ to perform better than non-ML. Furthermore, nlDAE converges faster than DAE, thus requiring less training data than DAE.

In Fig. \ref{fig:three graphs_mse}(c), Fig. \ref{fig:three graphs_ser}(c), and Fig. \ref{fig:three graphs_loc}(c), the impact of a noise-related parameter for each case study is illustrated. When the noise occurs according to a Bernoulli distribution in Fig. \ref{fig:three graphs_mse}(c), the performance of ML algorithms (both nlDAE and DAE) exhibits a concave behavior. This is because the variance of $\text{Ber}(p)$ is given by $p(1-p)$. Similar phenomenon is observed in Fig. \ref{fig:three graphs_loc}(c) because the Bernoulli event of NLoS constitutes a part of localization noise. As for non-ML, the performance worsens as the probability of noise occurrence increases in both cases. Fig. \ref{fig:three graphs_ser}(c) shows that the SER performance of nlDAE improves rapidly as the SNR increases. In all experiments, nlDAE gives superior performance than other schemes.

Thus far, the experiments have been conducted with a single hidden layer. Fig. \ref{fig:three graphs_mse}(d), Fig. \ref{fig:three graphs_ser}(d), and Fig. \ref{fig:three graphs_loc}(d) show the effect of the depth of the neural network. The performance of nlDAE is almost invariant, which suggests that nlDAE is not sensitive to the number of hidden layers. On the other hand, the performance of DAE worsens quickly as the depth increases owing to overfitting in two cases.

In summary, nlDAE outperforms DAE over the whole experiments. nlDAE is observed to be more efficient for the underlying use cases than DAE because it requires smaller latent space and less training data. Furthermore, nlDAE is more robust to the change of the parameters related to the design of the neural network, e.g., the network depth.

\section{Conclusion and Future Work}

We introduced a new denoiser framework based on the neural network, namely nlDAE.
This is a modification of DAE in that it learns the noise instead of the original data. The fundamental idea of nlDAE is that learning noise can provide a better performance depending on the stochastic characteristics (e.g., standard deviation) of the original data and noise.
We applied the proposed mechanism to the practical problems for IoT devices such as signal restoration, symbol demodulation, and precise localization.
The numerical results support that nlDAE is more efficient than DAE in terms of the required dimension of the latent space and the size of training dataset, thus rendering it more suitable for capability-constrained conditions.
Applicability of nlDAE to other domains, e.g., image inpainting, remains as a future work.
Furthermore, information theoretical criteria of decision making for the selection between or a combination of DAE and nlDAE is an interesting further research.

\bibliographystyle{IEEEtran}
\bibliography{references}
\end{document}